# AI-assisted workflow enables rapid, high-fidelity breast cancer clinical trial eligibility prescreening

Jacob T. Rosenthal[1], Emma Hahesy[2], Sulov Chalise[3], Menglei Zhu[3], Mert R. Sabuncu[4,5*], Lior Z. Braunstein[2*†], Anyi Li[6*†]

1. Weill Cornell Medicine/Rockefeller University/Memorial Sloan Kettering Cancer Center Tri-Institutional MD-PhD Program
2. Department of Radiation Oncology, Memorial Sloan Kettering Cancer Center
3. Department of Pathology and Laboratory Medicine, Memorial Sloan Kettering Cancer Center
4. School of Electrical and Computer Engineering, Cornell University and Cornell Tech
5. Department of Radiology, Weill Cornell Medicine
6. Department of Medical Physics, Memorial Sloan Kettering Cancer Center

*Equal contribution

†Correspondence to:
   Lior Braunstein, MD
   Department of Radiation Oncology
   Memorial Sloan Kettering Cancer Center
   BraunstL@mskcc.org

   Anyi Li, PhD
   Department of Medical Physics
   Memorial Sloan Kettering Cancer Center
   LiA5@mskcc.org

Word count (main text): 4509

# Abstract

Clinical trials play an important role in cancer care and research, yet participation rates remain low. We developed MSK-MATCH (Memorial Sloan Kettering Multi-Agent Trial Coordination Hub), an AI system for automated eligibility screening from clinical text. MSK-MATCH integrates a large language model with a curated oncology trial knowledge base and retrieval-augmented architecture providing explanations for all AI predictions grounded in source text. In a retrospective dataset of 88,518 clinical documents from 731 patients across six breast cancer trials, MSK-MATCH automatically resolved 61.9% of cases and triaged 38.1% for human review. This AI-assisted workflow achieved 98.6% accuracy, 98.4% sensitivity, and 98.7% specificity for patient-level eligibility classification, matching or exceeding performance of the human-only and AI-only comparisons. For the triaged cases requiring manual review, prepopulating eligibility screens with AI-generated explanations reduced screening time from 20 minutes to 43 seconds at an average cost of $0.96 per patient-trial pair.

# Introduction

Clinical trials are central to the advancement of cancer care and research, yet only 7–8% of eligible patients participate nationally[1]. This is despite studies showing that 44% of cancer patients have trials available to them at their treating institutions[2], and more than half of adults with cancer are likely to participate in a clinical trial when offered[3,4]. A key barrier is the complexity of eligibility determination, requiring information retrieval from longitudinal clinical documents spanning notes, imaging and pathology reports, and biomarkers.

Eligibility screening today is typically a labor-intensive manual process: screening a breast cancer patient for a single trial routinely takes longer than 30 minutes, requiring over an hour in more than 25% of cases, with overall cost estimates upwards of $330 per enrolled patient[5].

With systematic trial eligibility prescreening infeasible, current practice is mainly reliant on individual oncologists to stay informed about relevant trials and to advise patients of their options for clinical trial participation[6]. At large academic medical centers, which may have more than 1000 clinical trials recruiting at any given time, clinicians are unlikely to be able to stay up to date on all of them. Consequently, most oncology patients are never systematically screened for trials – many potential trial matches go undiscovered, patients are not offered the option of trial participation, and accrual rates remain low.

The emergence in recent years of large language models (LLMs), a type of AI model specialized for processing and generating textual data, has provided a powerful new tool for processing unstructured text including clinical notes and reports. Rather than being trained for a single task, current LLMs have general-purpose capabilities and show promise in many applications across cancer research and oncology[10–12]. Beyond standalone use, multiple LLMs can be orchestrated into multi-agent systems, with each agent performing a specialized role and communicating with others to perform complex, knowledge-intensive biomedical tasks[13–17].

Previous approaches for applying LLMs to automate clinical trial matching have been largely limited by the reliance on small datasets that are not representative of the complexity of longitudinal documentation in real-world data. For example, the 2018 n2c2 cohort selection challenge, with a maximum of 5 notes per patient[18], has been widely used for the development and evaluation of LLM-based eligibility screening systems, despite its limited representativeness of the scale of data seen in real clinical practice[19–22]. The performance of the Synapsis LLM for clinical trial eligibility assessment was evaluated on a set of 50 melanoma patients[23]. Other studies have relied on synthetic data, such as paragraph-length patient vignettes written by medical professionals[24–28], LLM-generated summaries of patient records[29], or even fully synthetic EHR data created by an LLM[30].

A second category of existing work adopts a two-step approach, using LLMs to extract prespecified sets of structured attributes, either biomarker based[31,32] or curated for trials of interest[23,33], before then applying rules-based logic to make a final eligibility determination. While this intermediate step could be automated by using LLMs to automatically extract relevant entities for trials[29,34], it is overall less scalable across diverse protocols than an end-to-end approach.

Some works have investigated using retrieval-augmentation to incorporate real-world clinical notes. For instance, the RECTIFIER tool showed promising performance when used for adjudicating 13 criteria in a single heart failure trial[35,36], but has not yet been extended to the oncology domain, where clinical trials are the most complex[37]. OncoLLM relies on an LLM fine-tuned on notes from a single institution and was evaluated on 98 patients, potentially limiting generalizability and adoption[38].

Building upon these previous efforts, the objective of our present work is to develop an AI-assisted clinical trial eligibility prescreening system which can be integrated with existing clinical workflows in a high-volume breast radiation oncology service at a large academic medical center to enable efficient, reliable, and systematic prescreening of new patients for relevant trials.

# Methods

## Cohort identification

We constructed a retrospective cohort of all patients referred for enrollment in an interventional clinical trial by the breast cancer radiation oncology service at Memorial Sloan Kettering Cancer Center (MSK) between 2016 and 2024. Once referred by a physician for potential enrollment in a particular trial, patients are then screened by trained staff at the hospital's centralized clinical trials office, who perform an in-depth review to determine the patient's eligibility status for the specific trial for which they were referred. These patient-level eligibility labels, representing the gold standard for trial eligibility determination in current standard of care at our institution, were used as ground truth. Criterion-level labels were not available, so for patients ultimately determined to be ineligible, we do not know the specific reason they were disqualified. After removing trials with fewer than 30 patients screened and fewer than two negative labels (i.e., patients determined to be ineligible after review), we were left with 731 patients spanning 6 trials (Table 1).

We obtained complete EHR data for each patient, including all available reports, notes, and other documents from all services. Documents created on or after the eligibility determination date were excluded to avoid information leakage. This yielded an average of 121 documents per patient, corresponding to 126,346 tokens as processed by the GPT-4o tokenizer[39] (Table 2).

This non-interventional research was approved by the MSK Institutional Review Board under protocol 16-114 and a waiver of informed consent was granted.

## Cross-trial negative labels

The data-generating process selected for patients presumed to be eligible, as patients were only referred for enrollment if the clinician believed they were very likely to qualify. Consequently, the original dataset was heavily skewed toward cases with a high pre-test probability of being eligible, and the negative labels it did contain likely represent the most difficult cases selected from a general population of largely ineligible patients, where the referring clinician was incorrect in their initial assessment. In reality, a clinician likely considers each new patient for all six of the possible trials, quickly ruling them out for the majority before identifying the trial(s) where they are likely to be eligible.

To mitigate class imbalance and capture a broader range of ineligible cases that better reflect real clinical decision-making, we therefore augmented our dataset with additional negative labels by leveraging mutually exclusive trials. Trials were classified as having metastatic disease as either an inclusion criterion (18-486, 22-259) or an exclusion criterion (16-323, 19-300, 19-410, 21-283), with patients eligible for a trial in one group being ineligible for the trials in the other group by definition. These cross-trial exclusions provided additional negative labels for the patients in our cohort at the same standard of evidence. After adding 1564 cross-trial negative labels to the original dataset of 731 patients, we obtained a total of 2295 {patient, trial} pairs (Table 1). The dataset was split into 50% training and 50% testing sets according to stratified random sampling to maintain label balance.

## Protocol data

The 6 clinical trials comprised a total of 73 inclusion criteria and 62 exclusion criteria (Table 3). The text for each criterion was taken verbatim from trial documentation, without any further processing.

In initial experimentation, we found that certain criteria were consistently difficult for the AI system to determine correctly. Some were written in such a way that they were always satisfied regardless of the patient, yet the AI model often made the incorrect determination (e.g., "bilateral breast cancer is permitted"). Others were judged by clinicians to be outside the scope of what could be automatically determined by AI and instead required clinical judgment by an oncologist on a case-by-case basis (e.g., whether a patient's disease is "technically amenable" to a particular therapy). This is reflective of the fact that while automated decision-making is desirable in many cases, there are certain instances where human expertise and input is required for high-stakes decisions. We flagged these criteria as "vacuous" and "requires human review" respectively, directing the AI to skip them and instead provide a predetermined answer ("met" or "unable to determine"). In total, we identified six vacuous criteria and 15 criteria requiring human review (Table 3, Table S1, Table S2).

## AI system design

Our objective was to design an AI system for automated trial eligibility determination that is effective and practical for use in real-world clinical workflows. To that end, the system must satisfy several criteria. First, predictions must be explainable and auditable to allow clinicians to verify the correctness of AI outputs, building trust and ensuring transparency. Secondly, the system should use general-purpose, commodity LLMs. With the rapid pace of technological advancement and the regular release of new models, we consider it essential to avoid being locked into a single model or provider, and to minimize the costs and complexity associated with training or fine-tuning LLMs on our data. Finally, the system must have the capacity to work with the scale and heterogeneity of real-world EHR data, with up to hundreds of notes per patient and dozens of criteria per trial, without requiring any manual curation.

The MSK-MATCH system comprises multiple LLM agents with distinct roles (**Figure 1**). We begin by creating expert agents emulating the composition of a multidisciplinary breast cancer care team, including a pathologist, radiologist, surgical oncologist, medical oncologist, radiation oncologist, and general medicine practitioner. Each agent is given access to a vector store containing clinical notes relevant to their specialty. Notes are assigned by matching their type to a predefined set of keywords and using prompts tailored to each specialty's domain. The generalist agent is assigned documents that do not match keywords for any of the other specialist agents. The set of expert agents is dynamically created for each patient based on their available documents, such that, for example, if a particular patient has no surgical notes, there will be no surgical oncology agent. In comparison to a single-expert design, the multi-expert framework showed comparable classification performance (Figure S1), but offered multiple domain-specific explanations which may improve overall interpretability. Notes from nursing, rehabilitation, and survivorship categories were excluded. Next, we create a trial coordinator agent responsible for routing each eligibility criterion to the appropriate expert agent(s) for evaluation. Finally, a principal investigator agent reviews the experts' explanations, resolves any discrepancies, and makes the final determination for each criterion (met, not met, or unable to determine). While this process could in principle be repeated in multiple rounds of deliberation, we found that a single pass was sufficient to achieve strong performance and did not explore further iterations. All prompts used are made available in the accompanying code release.

After all criteria have been assessed, a rules-based approach is applied to make the final patient-level eligibility determination. If any inclusion criteria are not met or any exclusion criteria are met, then the patient is classified as not eligible; otherwise, the patient is potentially eligible. Criteria labeled as "unable to determine" do not factor into this.

The workflow is defined as a graph structure using the LangGraph framework[40]. All agents used GPT-4o, version 2024-08-06[39]. The temperature was set to 0 to reduce creativity and encourage focused responses[41]. To protect patient data, all computation was conducted on an institutionally approved, HIPAA-compliant cloud computing instance on Microsoft Azure.

## Retrieval-augmented generation

A known shortcoming of LLMs is their tendency to generate responses that appear plausible yet contain references to nonexistent facts or sources: this phenomenon is known as hallucination[42–45]. Retrieval-augmented generation (RAG) is a technique commonly used to mitigate this problem by adding an additional step to LLM generation in which they first retrieve relevant snippets from a provided corpus of documents, which are then used to ground responses[46,47].

Retrieval augmentation was used to ground AI responses, allowing the LLM agents to search through a patient-specific vector store of clinical documents to identify relevant snippets before answering a query[48]. This facilitates transparency and auditability, improving response quality and increasing trust among users while reducing the risk of hallucination[46,47]. The notes are processed into a vector database as follows. First, each note is tokenized and split into chunks of 500 tokens, with an overlap of 50 tokens. Next, each chunk is passed through the Text-Embedding-3-Large model (OpenAI) to produce an embedding vector designed to capture the semantic meaning of the text within the chunk. Retrieval augmentation is then used to allow agents to access documents from the vector store during generation. When the LLM receives a query, the prompt is first converted into an embedding vector. A similarity search is then performed to retrieve the $k$ documents from the vector store whose embeddings are closest to the query embedding. In our implementation, we set $k = 10$ to select the top 10 most similar documents. These retrieved documents, containing the most relevant information, are then incorporated directly into the prompt, allowing the LLM to reference them when generating a response.

## Error analysis and knowledge base construction

In initial experiments, we observed overall poor performance. Experimentation demonstrated that individual errors could be rectified by modifying the system prompt, providing the LLMs with precise instructions to make the appropriate determination. However, we realized that hand-crafted prompting would not be a scalable solution. We therefore developed a human-in-the-loop framework to systematically analyze errors and solicit domain-specific feedback.

AI inference was run for the entire training set, and each misclassified case was individually reviewed by a clinical research coordinator (CRC) with experience screening patients for breast radiation oncology trials, who provided feedback in the form of clarification or instruction. The CRC was instructed to provide generally applicable feedback, rather than patient-specific. We collected all feedback in an oncology trial knowledge base (KB) (Figure 1b). During inference, the entire KB is concatenated together and injected directly into the prompts for each expert agent.

## AI-assisted workflow

We sought to develop a workflow leveraging the MSK-MATCH automated predictions to improve the efficiency of human review without compromising performance. During development, we observed that 1) AI predictions tend to have lower sensitivity than specificity, meaning that there are more false negatives than false positives; and 2) a predicted negative can be quickly confirmed by reviewing only the subset of criteria which have been flagged as disqualifying, whereas confirming a predicted positive requires checking every single criterion to verify eligibility.

To take advantage of these asymmetries, we created an AI-assisted workflow which conducts an automated first pass eligibility prediction and then triages a subset of the predicted negative cases for secondary human review. By sorting the predicted negatives according to the total number of disqualifying criteria (i.e., inclusion criteria not met, and exclusion criteria met), we are able to identify cases with only one or two disqualifying criteria. Limiting the review to these "almost matches" enables efficient use of humans' time on cases where their review has a higher likelihood of quickly identifying errors that change the final eligibility prediction. Furthermore, with the AI system having already extracted relevant information and prepopulated criterion-level explanations for each case, the secondary human review can be conducted faster.

## Experimental design

We conducted an experiment to compare three different workflows: a human only manual workflow, representing the current state of the art in clinical practice; a fully automated AI-only workflow; and a human-AI collaborative workflow in which the AI system prepopulates eligibility determinations and triages a subset for review by a human (Figure 2a).

First, AI inference was run on all n=1160 {patient, trial} pairs in the test set. Next, to measure the performance of the human-only workflow, we selected a random sample from the test set, stratified by protocol and class-balanced to create a sample of 5 positives and 5 negatives per protocol, for a total of 60. This subsampling was necessary due to the time-intensive nature of the manual workflow, which prohibited a manual evaluation of the entire test set. Samples were drawn from the original labels when possible and supplemented with cross-trial negatives only when there were fewer than 5 negative cases for a trial in the original dataset. The CRC was then asked to conduct a "gold-standard" exhaustive review for each of these cases, replicating the process they would use in the ordinary practice of their job, and free to use any resources apart from MSK-MATCH. The time duration was measured and recorded.

In the human-AI collaboration experimental arm, cases in the human-only baseline were excluded to ensure that none of the cases had been previously seen by the human curator, thereby avoiding any effects due to memorization or recall. This resulted in a total of n=1100 {patient, trial} pairs remaining. AI predictions were then sorted by the number of disqualifying criteria. The CRC was then asked to review all predicted ineligible cases with 1 or 2 disqualifying criteria. The time duration was measured and recorded. Performance statistics are calculated using the full set of 1100 predictions, including those which were triaged and did not receive human review, to capture the overall performance of the workflow.

## User interface

We developed a custom user interface to support both the human-in-the-loop error analysis and the human-AI collaborative screening workflow (Figure S2). The interface allows users to select a patient and protocol, then displays the patient-level result as well as individual criterion-level explanations. AI-generated explanations can be further inspected to view the verbatim snippets of text retrieved from the EHR data and used in the generation process. Dialogs for entering feedback allow for on-the-fly error analysis and correction. The web application is hosted behind the hospital's virtual private network (VPN) and password protected to ensure data security. The application runs on Kubernetes infrastructure and uses a Redis database on the backend.

## Statistical analysis

The primary outcome metrics were accuracy, sensitivity, and specificity. The Wilson score interval was used to compute confidence intervals for each metric. Due to our experimental setup, the experimental arms did not have an equal baseline prevalence of eligibility. To directly compare accuracy between groups with different prevalences, we created subsets of matched cases (Figure S3, Figure S4).

# Results

## Error analysis

Analysis of incorrect AI-only predictions identified four common failure modes: 1) gaps in domain knowledge, where the AI makes factual mistakes about breast cancer; 2) logical mistakes, where the AI misinterprets information or makes incorrect deductions; 3) incorrect handling of missing information, where the AI makes inappropriate inferences when elements are missing from a patient's EHR data; and 4) irrelevant criterion misclassification, where the model incorrectly applies or misinterprets a criterion that is only relevant in a specific context. Examples of each category are given in Table 4. Importantly, we did not observe any instances of hallucination, i.e. inclusion of factually incorrect information which is confabulated or otherwise not grounded in the underlying source documents. We conducted a single pass through the training set, collecting a total of 89 feedback elements capturing all error modes observed in the training set, forming a domain-specific oncology trial KB.

## Domain knowledge injection

We conducted an ablation study to assess the effect of including the domain knowledge in the KB, finding that injecting the KB directly into the prompts for each of the specialist agents increased accuracy from 64.1% (95% CI: 61.3-66.8%) to 87.5% (95% CI: 85.5-89.3%), increased sensitivity from 21.6% (95% CI: 17.6-26.2%) to 62.8% (95% CI: 57.6-67.7%), and increased specificity from 82.7% (95% CI: 79.9-85.1%) to 98.3% (95% CI: 97.1-99.0%) (Figure 3a). Addition of the KB into the prompt increased the number of tokens per LLM call, resulting in an increase in average cost per run of $0.58 (from $0.38 to $0.96) (Figure 3b). The KB only contained information obtained through review of errors in the training dataset, while performance was measured on the held-out testing dataset.

## Human-AI collaborative workflow

The human-AI collaborative workflow met or exceeded overall performance of the other workflows across all metrics, achieving accuracy of 98.6% (95% CI: 97.8-99.2%), sensitivity of 98.4% (95% CI: 96.4-99.3%), and specificity of 98.7% (95% CI: 97.7-99.3%) (Figure 2b). Full results are shown in Figure S5.

Applying a cutoff of two disqualifying criteria allowed us to capture nearly all the false negatives, while only requiring review of 38.1% of the AI predictions (Figure S6). The AI-assisted human workflow required an average of only 43 seconds to make a final eligibility determination, a 96.5% reduction from the manual baseline which required an average of 20 minutes and 25 seconds per case (Figure 2c). Confirmation of predicted positives, however, requires a full assessment of all criteria to identify any potential errors. Furthermore, the AI predictions have a higher positive predictive value than negative predictive value (94.04% vs. 85.84% respectively; Figure S5b) meaning that reviewing the predicted negatives is likely to be more fruitful in finding mistakes. Among the subset of cases that the human reviewed, the final classification performance was nearly perfect, with only a single false positive in 419 cases (Figure S5d).

# Discussion

MSK-MATCH demonstrates how a retrieval-grounded AI system can operationalize human-AI collaboration for high-stakes clinical decision-making. Applied here to breast cancer trial eligibility prescreening, the system achieves clinical-grade accuracy while dramatically reducing manual human effort and cost. We developed and evaluated MSK-MATCH using a large retrospective dataset of more than 88,000 clinical documents from 731 breast cancer patients across 6 interventional breast cancer trials. In comparison to related works, our evaluation is conducted using a dataset which more faithfully captures the challenges of a busy clinical service at a large academic cancer center, with full longitudinal medical records, no synthetic data, and multiple clinical trials (Table S3).

Because eligibility determinations produced by our screening system are intended to be used as inputs for downstream steps in the clinical workflow, it is important for real-world utility that the performance is sufficiently good to not overwhelm the system with false positives. The positive predictive value (PPV) of our human-AI collaborative workflow, as measured on the test set, is not significantly different from the PPV of the current state-of-the-art as measured by the distribution of labels in our original dataset (96.9% vs. 95.2%, two-tailed normal test for proportions, P=0.196). Therefore, the use of the AI-assisted system is not expected to result in substantially more false positives than current practice, further highlighting its readiness for integration with existing workflows in real-world clinical use. With an average cost of less than one dollar to evaluate a patient for a single trial, implementation of broad prescreening programs for trial eligibility is feasible, and there is opportunity for future work to investigate reducing costs further.

The paradigm of human-AI collaboration that first uses AI to autonomously and rapidly produce predictions and explanations and then incorporates human experts to verify and refine these outputs has shown success in other medical applications such as generating radiology reports[49], giving feedback to surgical trainees[50], and responding to patient messages[51]. In addition to boosting predictive accuracy, such approaches also help build trust among users, smoothing the path toward widespread adoption[52]. However, there have been concerns that reliance on humans to review LLM-generated text could result in added cognitive burden due to the tedious nature of the work, possibly inducing cognitive biases (e.g. automation bias)[53]. To the contrary, we showed that our proposed workflow can largely automate what is today a highly labor-intensive and tedious manual process, with the net result of decreased cognitive load and increased efficiency. Future work to further optimize workflows and user interfaces may be able to further improve human-AI collaboration.

MSK-MATCH is designed to incorporate multiple layers of explainability and transparency with increasing levels of granularity, thereby facilitating the verification of AI outputs. At the highest level, the final eligibility determination is highlighted along with the counts of inclusion and exclusion criteria by status. Eligibility status can then be broken down by criterion-level labels, which is especially helpful in enabling users to quickly identify which specific criteria disqualify a patient from a given trial. For each criterion, integrative explanations synthesize multiple sources of evidence to provide a narrative justification for the AI model's determination with references to source documents as evidence. Finally, citations in explanations can be traced back to verbatim snippets of text from original source documents in the medical record. Together, these mechanisms help build trust with clinical users while reducing the occurrence of AI model hallucinations or factual errors. This trust translates into practical gains, with the CRC taking less than one minute on average to review AI predictions in our experiment, suggesting that they are reviewing only a subset of criteria for correctness and trusting the majority of the AI outputs – leading to substantial efficiency gains.

How best to leverage general-purpose LLMs in highly specialized medical domains remains an active area of inquiry. Previous works have built domain-specific models by conducting additional fine-tuning or retraining entirely from scratch with a narrowly curated dataset of clinically relevant text, such as EHR data[38,54–58]; however, such domain adaptation can be costly and logistically difficult. Other work has cast doubt on the value of specialized models, positing that general-purpose LLMs can achieve comparable performance even on narrow medical tasks[59,60]. Our work takes a middle path, using a commodity LLM while imbuing it with specialized domain knowledge through its prompt, achieving strong performance in complex information extraction and reasoning without the need to invest in model training or fine-tuning. This approach is pragmatic, allowing the base LLM to be swapped for newer versions as vendors release them. Furthermore, the KB can continue to be dynamically updated while the system is deployed in clinical practice, allowing the AI system to continually improve and adapt in response to user feedback[61]. In summary, we emphasize the importance of incorporating specialized domain knowledge to unlock the full potential of generalist AI models for oncology applications, noting that this can be done through prompting strategies without requiring fine-tuning.

Our study does have limitations. Firstly, while our dataset of real longitudinal patient records and gold-standard eligibility labels allows us to comprehensively evaluate our proposed workflow in scenarios similar to how it would be used in real clinical practice, it still represents a single service at a single academic center and may not capture differences in population and practice patterns in other cancer types, institutions, or regions. Secondly, we did not systematically evaluate the performance of different LLMs, deciding to instead focus on human-AI collaboration and clinical workflow integration. Further work should assess how changing the base LLM from GPT-4o to other models, either from commercial vendors or local LLMs from the open-source community, could affect performance and cost. Additionally, our method of directly injecting the entire KB into the LLM prompts, while effective, did increase inference costs. A targeted approach to retrieve relevant information from the KB could preserve performance gains while reducing prompt lengths, improving scalability and efficiency.

# Conclusion

We developed a workflow that effectively leverages AI for automated first-pass eligibility screening for breast cancer clinical trials, prepopulating explanations and identifying relevant documents while triaging only a subset of cases for secondary human review. By reducing eligibility prescreening as a barrier to trial entry, more patients can be offered the option of clinical trial participation, increasing access to experimental therapies and accelerating the pace of clinical oncology research to advance the standard of care for future patients. MSK-MATCH demonstrates a pragmatic and effective approach to harnessing AI in clinical workflows which could serve as a blueprint for other applications beyond clinical trial eligibility screening.

# Acknowledgements

J.T.R. was supported by a Medical Scientist Training Program grant from the National Institute of General Medical Sciences of the National Institutes of Health under award number T32GM152349 to the Weill Cornell/Rockefeller/Sloan Kettering Tri-Institutional MD-PhD Program. This study was supported in part by the Achar Meyohas Family and NIH/NCI Cancer Center Support Grant No. P30CA008748.

# Author contributions

J.T.R. contributed project conceptualization, methods and experimental design, implementation, and analysis of results. E.H. reviewed AI predictions and contributed to experimental design. S.C. and M.Z. contributed to methods and experimental design and analysis of results. M.R.S, L.Z.B, and A.L. contributed to project conceptualization, methods and experimental design, analysis of results, and project oversight. All authors contributed to writing and review of the manuscript.

## Data availability

The patient-level longitudinal medical record data cannot be shared publicly due to patient privacy protections. Eligibility criteria for the protocols used in this study are publicly available at https://github.com/msk-mph/msk-match.

## Code availability

Code to reproduce the methods is publicly available at https://github.com/msk-mph/msk-match.

# Figures

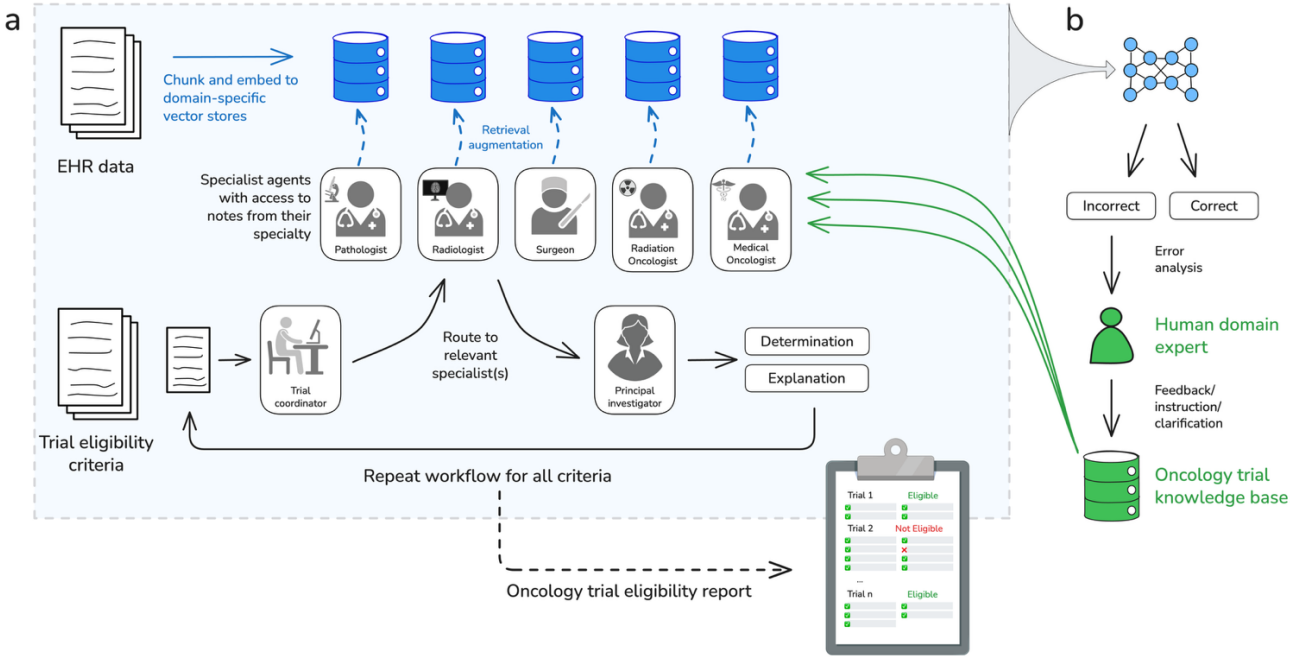

**Figure 1**. Schematic diagram of the MSK-MATCH system and human-in-the-loop knowledge base curation. **a)** Clinical documents from EHR are categorized by domain, then split into chunks and embedded to create vector stores (solid blue arrow). Each trial eligibility criterion is sent through the multi-expert workflow (solid black arrows). Expert agents have access to the vector store of documents from their specialty for retrieval-augmented generation (dashed blue arrows). After all criteria have been assessed, a final report with a patient-level eligibility determination and explanations for each criterion is produced (dashed black arrow). **b)** The AI system was evaluated on all {patient, trial} pairs in the training dataset. Every incorrect prediction was manually reviewed by a clinical research coordinator, and feedback was given in the form of instruction for the LLMs on how to avoid making the same error in the future. All feedback was collected into an oncology trial knowledge base. When evaluating on the test dataset, the knowledge base is injected directly into the prompts of the expert agents to guide their outputs (green arrows). Figure created with icons from Biorender.com.

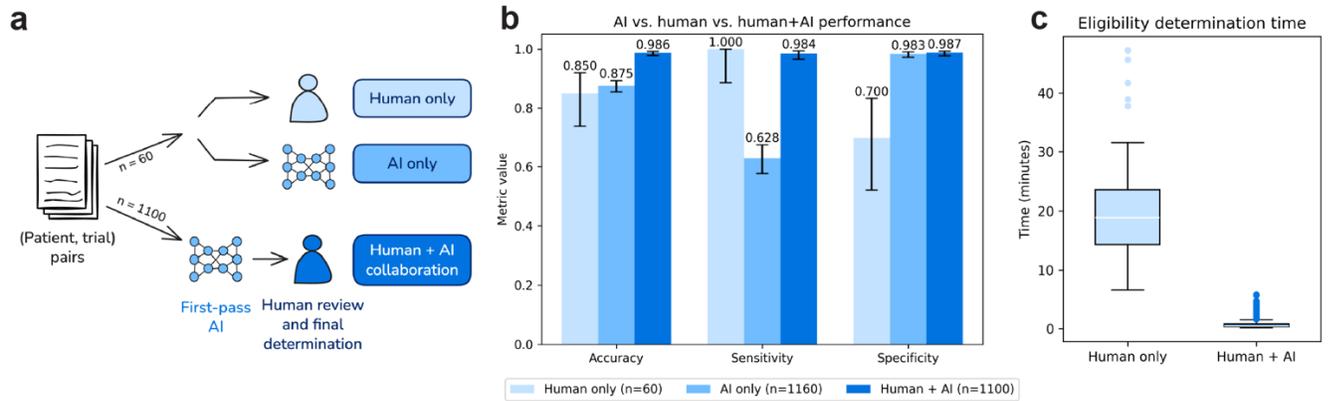

**Figure 2**. Experimental design and overall results. **a)** Schematic of the experimental design. All {patient, trial} pairs are evaluated by the MSK-MATCH AI system ("AI alone", n=1160). All cases are also assigned to human review, either with the AI-assisted triage and workflow ("Human + AI", n=1100) or with the unassisted manual workflow ("Human alone", n=60). **b)** Comparison of performance metrics on the test set, by experimental group. **c)** Comparison of the time duration required to complete eligibility screening for a single {patient, trial} pair, with and without AI-assistance.

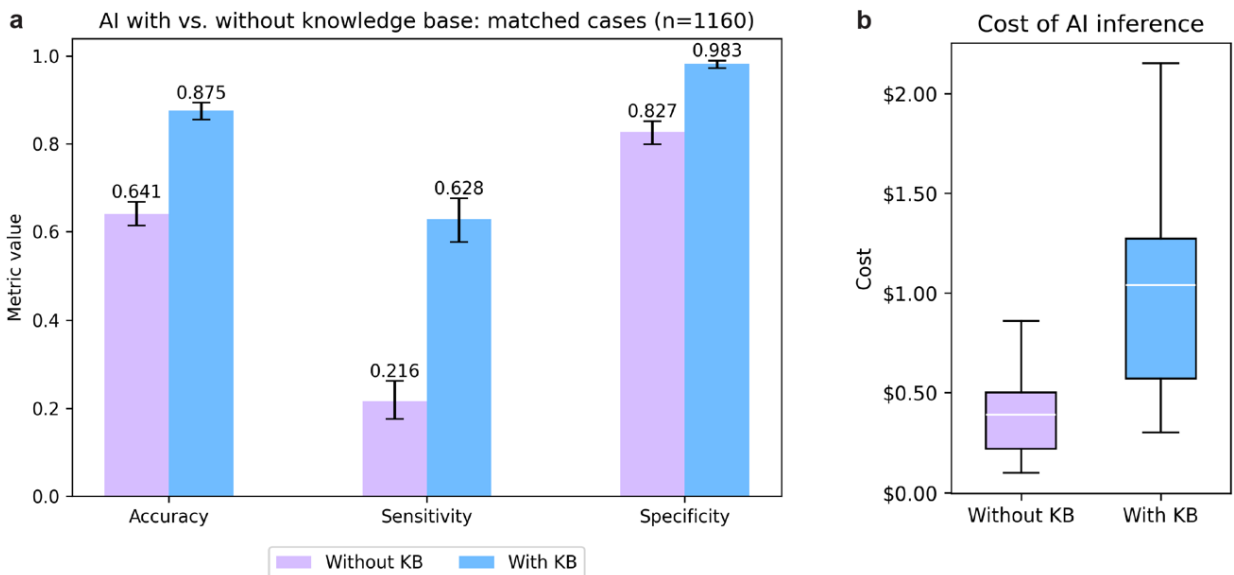

**Figure 3**. Effect of including the oncology trial knowledge base (KB). **a)** Comparison of performance metrics on the test set, with and without inclusion of the KB. **b)** Comparison of cost per {patient, trial} pair, with and without inclusion of the KB.

# Tables

**Table 1**. Dataset summary.

| Protocol | Eligible | Not Eligible (Original) | Not Eligible (Cross-Trial) | Total |
|---|---|---|---|---|
| **16-323** | 387 | 13 | 86 | **486** |
| **18-486** | 56 | 5 | 610 | **671** |
| **19-300** | 167 | 5 | 86 | **258** |
| **19-410** | 29 | 3 | 86 | **118** |
| **21-283** | 27 | 6 | 86 | **119** |
| **22-259** | 30 | 3 | 610 | **643** |
| **Total** | **696** | **35** | **1564** | **2295** |

**Table 2**. Summary statistics of clinical documents. Token count is calculated using the GPT-4o tokenizer[39].

| Statistic | Documents | Tokens |
|---|---|---|
| **Mean** | 121.09 | 126345.88 |
| **Total** | 88518 | 92358837 |

**Table 3**. Details of protocols included in this study. Inclusion and exclusion criteria are drawn directly from clinical trial documentation. A subset of criteria is labeled as "vacuous" or "requires human review" based on manual inspection (see Table S1 and Table S2). NCT ID: National Clinical Trial Identifier.

| Protocol | NCT ID | Inclusion Criteria | Exclusion Criteria | Total | Vacuous | Requires Human Review |
|---|---|---|---|---|---|---|
| **16-323** | NCT02603341 | 12 | 5 | 17 | 3 | 2 |
| **18-486** | NCT03808337 | 12 | 6 | 18 | 0 | 2 |
| **19-300** | NCT04084730 | 9 | 15 | 24 | 0 | 2 |
| **19-410** | NCT03488693 | 16 | 10 | 26 | 2 | 4 |
| **21-283** | NCT04852887 | 17 | 22 | 39 | 1 | 3 |
| **22-259** | NCT05534438 | 7 | 4 | 11 | 0 | 2 |
| **Total** | | **73** | **62** | **135** | **6** | **15** |

**Table 4**. Representative examples of AI error modes.

| Error Type | Criterion | AI Output: Explanation and Determination | Error Explanation |
|---|---|---|---|
| **Gaps in Domain Knowledge** | Inclusion: Metastatic disease detected on imaging and histologically confirmed. | "The report from the FDG PET/CT scan on [date] indicates a hypermetabolic left breast neoplasm metastatic to the left axilla, which is consistent with metastatic disease."<br><br>Determination: Met | Spread to axillary nodes, although described as "metastatic" in the imaging reports, does not count as "metastatic disease" for the purposes of this exclusion criterion. |
| | Exclusion: History of cosmetic or reconstructive breast surgery. | "The patient has a history of breast surgery, specifically a partial mastectomy, as indicated in the follow-up report. This is considered a form of reconstructive breast surgery. Therefore, the patient has a history of cosmetic or reconstructive breast surgery."<br><br>Determination: Met | Mastectomies and lumpectomies are not considered cosmetic or reconstructive breast surgery. |
| **Logical Mistakes** | Inclusion: The patient must be ≥50 years and <70 years of age. | "The criterion requires a patient to be a female between the ages of 50 and 69. The progress notes from [date] state that the patient is a 65-year-old female. Therefore, the criterion is Met."<br><br>Determination: Not Met | Final determination is incongruent with explanation. |
| | Exclusion: Prior radiation therapy to the ipsilateral or contralateral breast or thorax. | "The treatment plan from [date] mentions radiation therapy to the breast and axilla. Therefore, the exclusion criterion is met."<br><br>Determination: Met | Discussion of a planned future treatment, mistaken for treatment history. |
| | Exclusion: Dermatomyositis with a CPK level above normal or with an active skin rash or scleroderma. | "The patient has a rash, as noted in both the initial consult and follow-up visits. Thus, the patient meets this criterion"<br><br>Determination: Met | Criterion as written in protocol is logically ambiguous. Logic of the form "A with B or C or D" interpreted by LLM as "(A with B) or C or D" when it likely was intended to be "(A with B or C) or D". |
| **Incorrect Handling of Missing Information** | Inclusion: ECOG Performance Status of 0 or 1. | "The document does not mention ECOG performance status."<br><br>Determination: Not Met | If a criterion requires specific concrete data elements such as ECOG status, BMI, or age, and the data element is not explicitly mentioned in the patient's chart, the criterion should be marked as 'unable to determine' and flagged for human review. |
| | Inclusion: Patients who are HIV positive are eligible, provided they are under treatment with highly active antiretroviral therapy and have a CD4 count ≥ 200 cells/microliter within 180 days prior to registration. | "The patient has no mention of HIV in their chart. Without explicit confirmation of their HIV status, I am unable to determine whether they 'meet' the criterion."<br><br>Determination: Not Met | If there is no mention of HIV status or CD4 count tests, then it is reasonable to assume that the patient is not HIV positive. |
| **Irrelevant Criterion Misclassification** | Inclusion criteria for intermediate risk sub-study: Invasive lobular carcinoma. | "The criterion specifies that a patient must have biopsy-proven invasive lobular carcinoma to be eligible for the intermediate risk sub-study group. According to the surgical pathology on [date], the patient has a diagnosis of invasive ductal carcinoma. Therefore, she does not meet the eligibility criterion."<br><br>Determination: Not Met | Criteria describing sub-studies should be marked as "Met" or "Unable to determine" if not relevant to the patient, so as to not adversely affect their overall eligibility assessment for the main study. |
| | Inclusion: Patients with brain metastases are eligible if these lesions have been treated prior to enrollment. | "Radiology report from [date] shows no evidence of metastatic disease."<br><br>Determination: Not Met | For criteria in the form 'patients with X are eligible if Y,' showing that a patient does not have X is sufficient to answer the criterion. |

# Supplementary Information

## Supplementary Figures

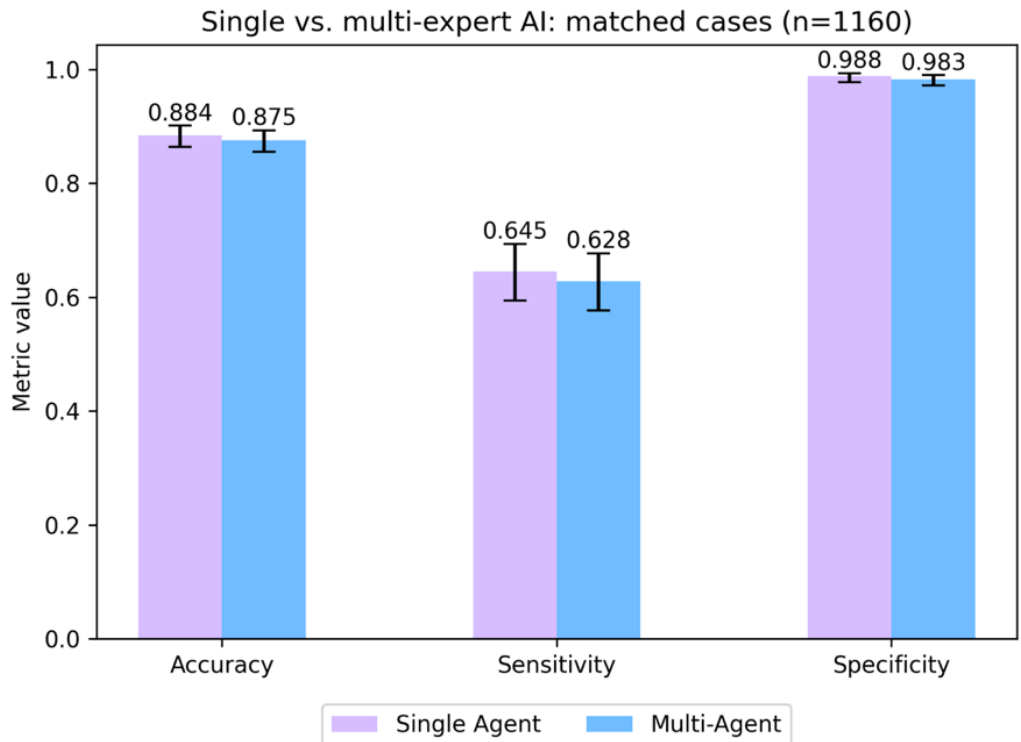

**Supplementary Figure S1**. Comparison of performance metrics on the test set, single-expert vs. multi-expert system architecture.

**Supplementary Figure S2**. MSK-MATCH web application human-AI interface. **a)** The top row has controls for selecting protocol and patient-specific medical record number (MRN). When data are loaded for the selected {patient, protocol} pair, the overall eligibility status is shown, along with counts of criteria by status (qualifying, disqualifying, and unable to determine). **b)** Example of interface for one criterion. On the left column, the criterion label is colored by status to provide a visual cue for users scrolling through. The second column gives the verbatim text of the criterion. The third column displays the AI-generated explanations, one for each of the expert agents consulted. In the final column, a dropdown menu allows the user to override the AI-generated explanation based on their human expert judgement. In that case, a modal dialog box will prompt the user to provide feedback or explanation of their decision, which is then added to the oncology trial knowledge base. **c)** When the "inspect evidence" button is clicked, users are directed to a separate page (depicted here in an inset) where they can view the verbatim snippets from the clinical documents which were retrieved from the vector stores and used by the expert agent. Each snippet is labeled with the note type and date to facilitate users being able to quickly navigate to it in the external EHR system for further inspection and verification. Dates and other identifying information have been redacted to protect privacy.

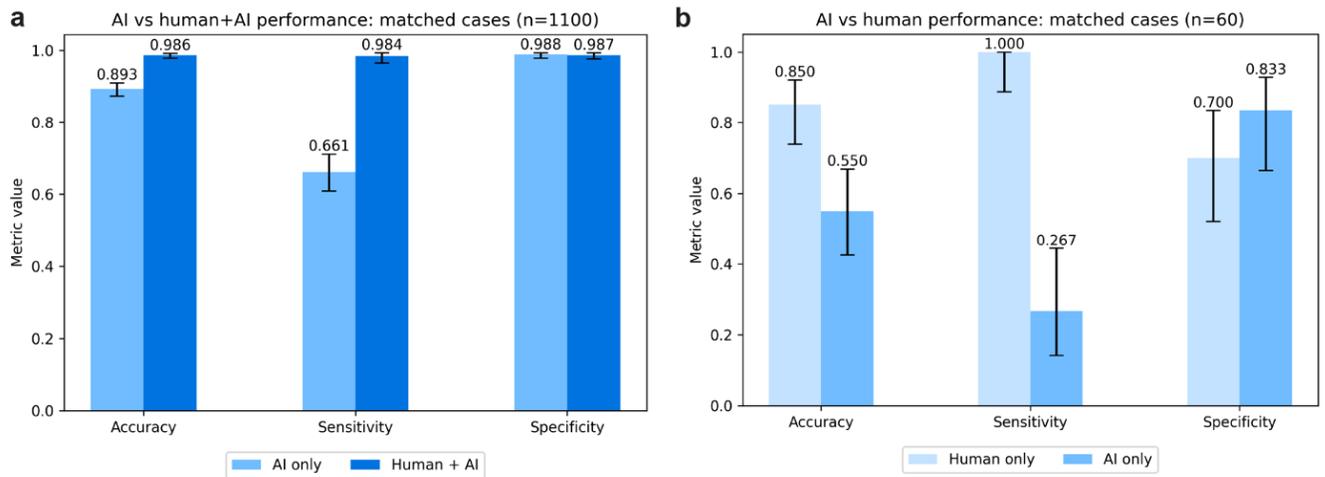

**Supplementary Figure S3**. Performance on matched cases in the test set. **a)** AI only vs. human-AI collaborative workflow. **b)** Human only vs. AI only.

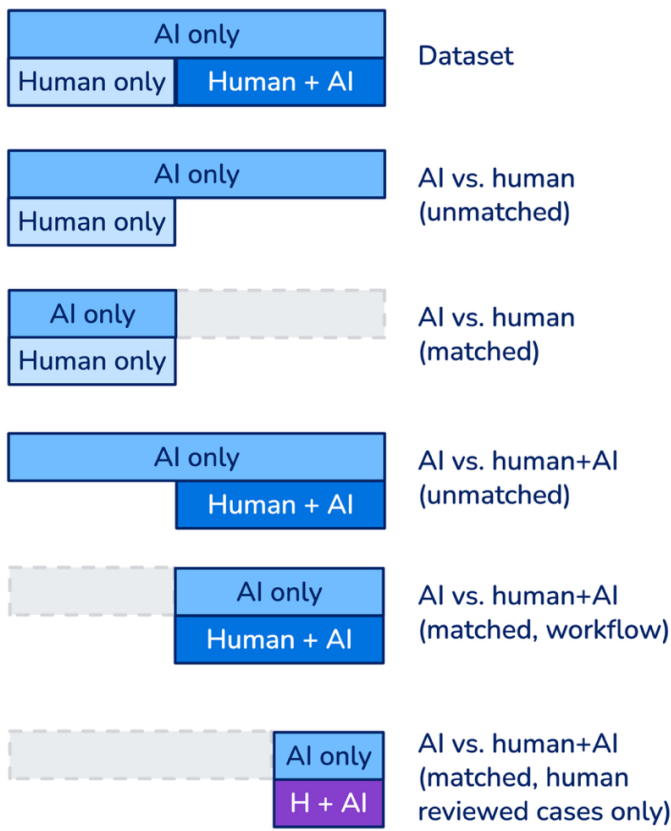

**Supplementary Figure S4**. Schematic diagram of dataset resulting from experimental design shown in Figure 2a.

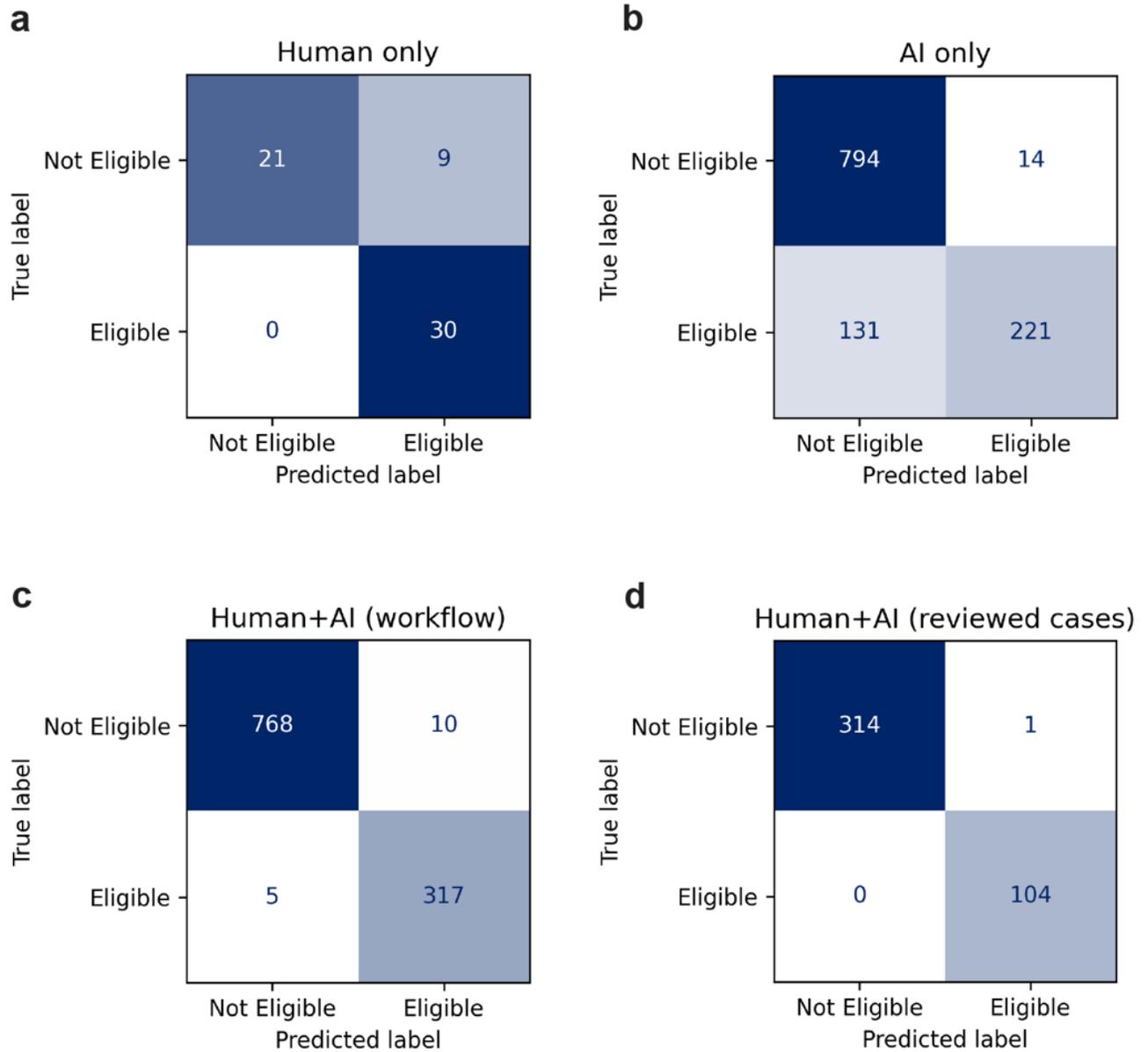

**Supplementary Figure S5**. Detailed results of the classification performance of four different conditions on the test set. **a)** Human-only workflow, unassisted. **b)** AI predictions only, with no human review. **c)** Human-AI collaborative workflow, with triage and human review of a subset of AI predictions. **d)** Performance on the subset of cases from (c) that were reviewed by a human.

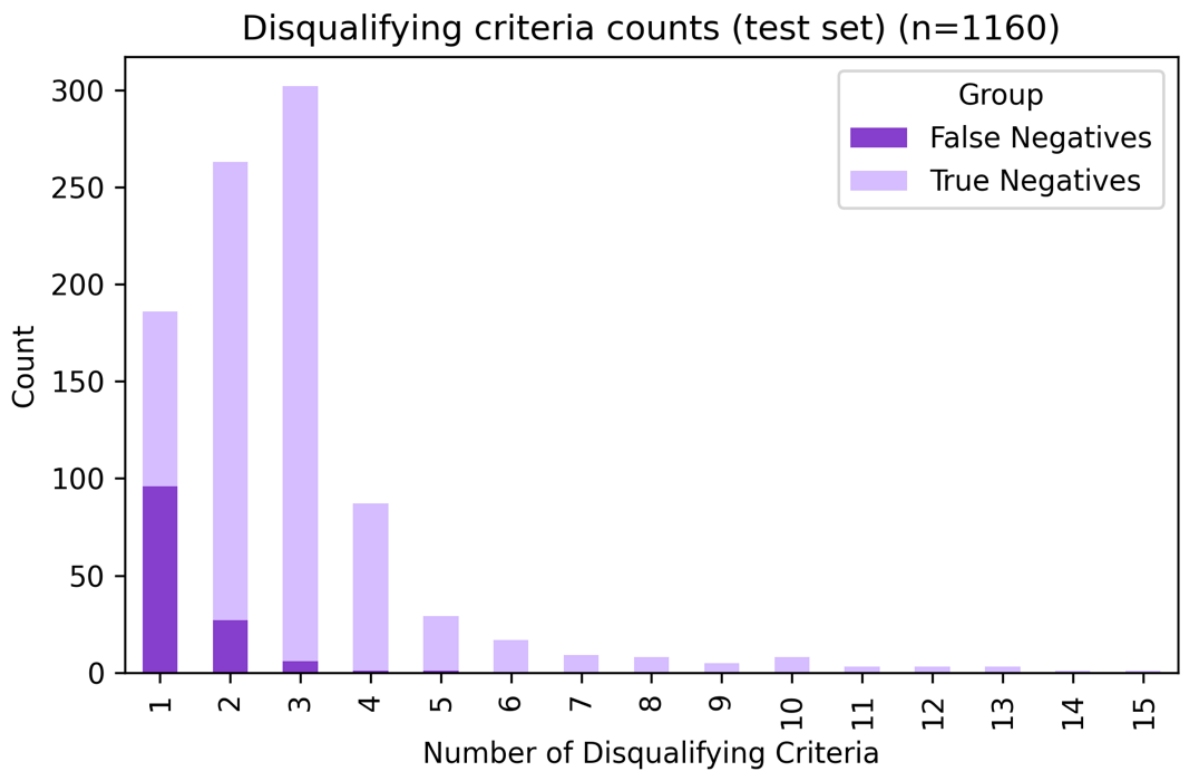

**Supplementary Figure S6**. Distribution of the number of disqualifying criteria among predicted negatives in the test set. A disqualifying criterion is an inclusion criterion that is not met or an exclusion criterion that is met. Colors indicate the counts of true negatives and false negatives.

## Supplementary Tables

**Supplementary Table S1**. Text of criteria identified as vacuous.

| Criterion | Protocol |
|---|---|
| For patients who have undergone lumpectomy, any type of mastectomy and any type of reconstruction (including no reconstruction) are allowed. Metallic components of some tissue expanders may complicate delivery of proton therapy; any concerns should be discussed with the Breast Committee Study Chairs prior to registration. | 16-323 |
| For patients who have undergone lumpectomy, there are no breast size limitations. | 16-323 |
| Bilateral breast cancer is permitted. Patients receiving treatment to both breasts for bilateral breast cancer will be stratified as left-sided. | 16-323 |
| Patients may or may not have had adjuvant chemotherapy. | 19-410 |
| Patients with T3N0 disease are eligible. | 19-410 |
| The trial is open to female and male patients. | 21-283 |

**Supplementary Table S2**. Text of criteria identified as requiring clinician input.

| Criterion | Protocol |
|---|---|
| Confirmation that the patient's health insurance will pay for the treatment in this study (patients may still be responsible for some costs, such as co-pays and deductibles). If the patient's insurance will not cover a specific treatment in this study and the patient still wants to participate, confirmation that the patient would be responsible for paying for any treatment received. | 16-323 |
| The patient must provide study-specific informed consent prior to study entry. | 16-323 |
| Able to provide informed consent. | 18-486 |
| Patients whose entry to the trial will cause unacceptable clinical delays in their planned management. | 18-486 |
| Written informed consent obtained from subject and ability to comply with the requirements of the study. | 19-300 |
| For female subjects of childbearing potential, patient is willing to use 2 methods of birth control or be surgically sterile or abstain from heterosexual activity for the duration of study participation. Note: Should a woman become pregnant while participating on study, she should inform the treating physician immediately. | 19-300 |
| Patient consent must be appropriately obtained in accordance with applicable loca`l and regulatory requirements. Each patient must sign a consent form prior to enrollment in the trial to document their willingness to participate. A similar process must be followed for sites outside of Canada as per their respective cooperative group's procedures. | 19-410 |
| Patients must be accessible for treatment and follow-up. Investigators must assure themselves that patients randomized on this trial will be available for complete documentation of the treatment, adverse events, and follow-up. | 19-410 |
| Patients must have had endocrine therapy initiated or planned for ≥ 5 years. Premenopausal women will receive ovarian ablation plus aromatase inhibitor therapy or tamoxifen if adjuvant chemotherapy was not administered. For all patients, endocrine therapy can be given concurrently or following RT. | 19-410 |
| Has the patient seen their Medical Oncologist? | 19-410 |
| Women of childbearing potential must have agreed to use an effective contraceptive method. A woman is considered to be of 'childbearing potential' if she has had menses at any time in the | 19-410 |
| preceding 12 consecutive months. In addition to routine contraceptive methods, 'effective contraception' also includes heterosexual celibacy and surgery intended to prevent pregnancy (or with a side-effect of pregnancy prevention) defined as a hysterectomy, bilateral oophorectomy or bilateral tubal ligation, or vasectomy/vasectomized partner. However, if at any point a previously celibate patient chooses to become heterosexually active during the time period for use of contraceptive measures outlined in the protocol, she is responsible for beginning contraceptive measures. Women of childbearing potential will have a pregnancy test to determine eligibility as part of the Pre-Study Evaluation (see Section 4.0); this may include an ultrasound to rule-out pregnancy if a false-positive is suspected. For example, when beta-human chorionic gonadotropin is high and partner is vasectomized, it may be associated with tumour production of hCG, as seen with some cancers. Patient will be considered eligible if an ultrasound is negative for pregnancy. | |
| The patient or a legally authorized representative must provide study-specific informed consent prior to pre entry /step 1 and, for patients treated in the U.S., authorization permitting release of personal health information. | 21-283 |
| Patients must be intending to take endocrine therapy for a minimum 5 years duration (tamoxifen or aromatase inhibitor). The specific regimen of endocrine therapy is at the treating physician's discretion. | 21-283 |
| Willing and able to provide informed consent. | 22-259 |
| Consented to 12-245. | 22-259 |

**Supplementary Table S3**: Comparison of datasets used to evaluate performance in related works. ◇: pre-specified set of criterion-level attributes; ◈: single trial; ◆: multiple trials.

| Reference | Number of patients | Documents per patient | Real EHR data | Clinical trial data |
|---|---|---|---|---|
| 2018 n2c2 cohort selection challenge[18] (used in [19–22]) | 288 | 2-5 | ✓ | ◇ |
| TREC 2021 Clinical Trials Track[26] | 75 | 1 | ✗ | ◆ |
| TREC 2022 Clinical Trials Track[25] | 50 | 1 | ✗ | ◆ |
| Ferber et al.[30] | 51 | 1 | ✗ | ◆ |
| TrialGPT[27] | 183 | 1 | ✗ | ◆ |
| Trial-LLAMA[28] | 313 | 1 | ✗ | ◆ |
| Synapsis[23] | 50 | ? | ✓ | ◇ |
| Parikh et al.[33] | 74 | ? | ✓ | ◇ |
| MatchMiner-AI[29] | 13086 | 1[†] | ✗[†] | ◆ |
| OncoLLM[38] | 98 | 74 | ✓ | ◆ |
| RECTIFIER[35] | 1509 | 127* | ✓ | ◈ |
| **MSK-MATCH (ours)** | **731** | **121** | ✓ | ◆ |

[†] Synthetic summaries derived from real EHR data
* Estimated using values from Figure S2 C.